\documentclass[lettersize,journal]{IEEEtran}
\usepackage{amsmath,amsfonts}
\usepackage{algorithmic}
\usepackage{array}
\usepackage[caption=false,font=normalsize,labelfont=sf,textfont=sf]{subfig}
\usepackage{textcomp}
\usepackage{stfloats}
\usepackage{url}
\usepackage{verbatim}
\usepackage{graphicx}
\usepackage{arydshln}
\def\BibTeX{{\rm B\kern-.05em{\sc i\kern-.025em b}\kern-.08em
    T\kern-.1667em\lower.7ex\hbox{E}\kern-.125emX}}
\usepackage{balance}
\begin{document}
\title{CellMix: A General Instance Relationship based Method for Data Augmentation Towards Pathology Image Classification}
\author{Tianyi Zhang, Zhiling Yan,
        Chunhui Li, Nan Ying,
        Yanli Lei, Yunlu Feng, Yu Zhao, Guanglei Zhang, ~\IEEEmembership{Member, IEEE}
\thanks{This work was partially supported by the National Natural Science Foundation of China (No. 62271023), the Beijing Natural Science Foundation (No. 7202102), the Fundamental Research Funds for Central Universities, and the National High Level Hospital Clinical Research Funding (No. 2022-PUMCH-177).

Tianyi Zhang, Zhiling Yan and Chunhui Li contributed equally to this work. Corresponding Author: Guanglei Zhang (e-mail: guangleizhang@buaa.edu.cn) 

Tianyi Zhang, Nan Ying, Yanli Lei,  and Guanglei Zhang are with the Beijing Advanced Innovation Center for Biomedical Engineering, School of Biological Science and Medical Engineering, Beihang University, Beijing, 100191, China (e-mails: {zhangtianyi, 20101014, 20373207,  guangleizhang}@buaa.edu.cn)

Zhiling Yan is with the School of Biological Sciences, Nanyang Technological University, Singapore, 639798, Singapore (e-mail: zhilingyan724@outlook.com)

Chunhui Li is with the School of Artificial Intelligence, Nanjing University, Nanjing, 210093, China (e-mail: lich@smail.nju.edu.cn)

Yunlu Feng  is with the Department of Gastroenterology, Peking Union Medical College Hospital, Beijing, 100006, China (e-mail: yunluf@icloud.com)

Yu Zhao is with the Department of Pathology, Peking Union Medical College Hospital, Beijing, 100006, China (e-mail: rain986532@126.com)

}}


\maketitle

\begin{abstract}
In pathology image analysis, obtaining and maintaining high-quality annotated samples is an extremely labor-intensive task. To overcome this challenge, mixing-based methods have emerged as effective alternatives to traditional preprocessing data augmentation techniques. Nonetheless, these methods fail to fully consider the unique features of pathology images, such as local specificity, global distribution, and inner/outer-sample instance relationships. To better comprehend these characteristics and create valuable pseudo samples, we propose the CellMix framework, which employs a novel distribution-oriented in-place shuffle approach. By dividing images into patches based on the granularity of pathology instances and shuffling them within the same batch, the absolute relationships between instances can be effectively preserved when generating new samples. Moreover, we develop a curriculum learning-inspired, loss-driven strategy to handle perturbations and distribution-related noise during training, enabling the model to adaptively fit the augmented data. Our experiments in pathology image classification tasks demonstrate state-of-the-art (SOTA) performance on 7 distinct datasets. This innovative instance relationship-centered method has the potential to inform general data augmentation approaches for pathology image classification. The associated codes are available at https://github.com/sagizty/CellMix.
\end{abstract}


\section{Introduction}

\begin{figure}
  \centering
  \includegraphics[width=3in]{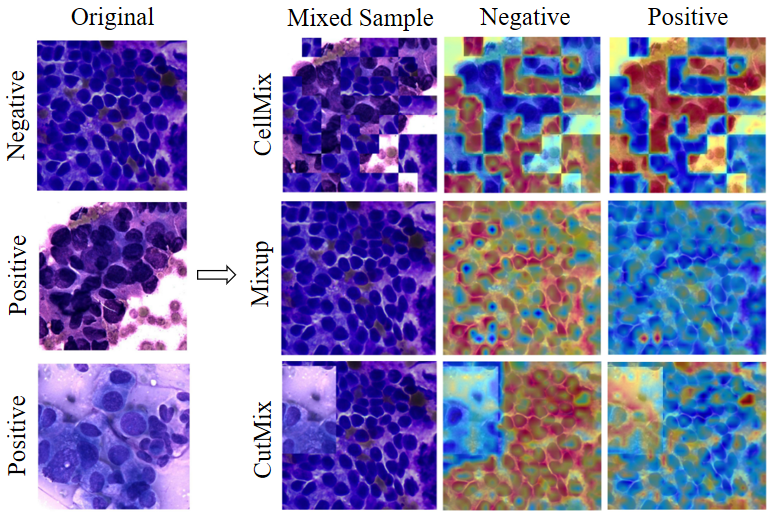}
  \caption{Visualization of an augmented example. Compared with other mixing-based methods, CellMix accurately identifies the boundary of patches and instances. It also tells the differences between negative and positive.
}
  \label{fig:CAM_augmented}
\end{figure}
\IEEEPARstart{D}{eep} neural networks (DNNs) have shown impressive performance in pathology image analysis. However, this success crucially relies on the availability and quality of annotated pathology images, which are very difficult to collect and need lots of human effort. To alleviate this problem, data augmentation, which enlarges the amount and diversity of the labeled samples with label-preserving transformations, is a feasible method to improve the generalization of DNNs.

Data augmentation has been extensively established in the traditional computer vision field. Beyond the traditional preprocess data augmentation methods (including: rotate, flip, etc.), the mixing-based approach is one of the most representative methods.
Mixing-based methods are domain-agnostic data augmentation techniques first proposed by \cite{zhang2017mixup}. It devises a strategy that mixes some sub-regions or features from different categories of images, as new pseudo samples. Specifically, Mixup \cite{zhang2017mixup}, Cutout \cite{devries2017improved}, and CutMix \cite{yun2019cutmix} apply the mixing process at the instance level. 
While Patchup \cite{faramarzi2020patchup} trims the hidden layers and mixes the image at the feature level. 
PuzzleMix \cite{kim2020puzzle} and SaliencyMix \cite{uddin2020saliencymix} first devise a strategy to extract the salient instance of the original images, then establish the mixing process.

Specifically, in the pathology field, two mainstream data augmentation methods are widely adopted, including Mixing-based methods and GAN-based methods. For Mixing-based methods, MixPatch \cite{park2022mixpatch} trains a CNN-based histopathology patch-level classifier to randomly mix
patches from different images.
Stain mix-up \cite{chang2021stain} 
aim to address stain color variants to perform a stain mix-up process.  InsMix \cite{lin2022insmix} combines instances with similar morphology characteristics and shuffles background patches.
Scorenet \cite{stegmuller2023scorenet} pastes one patch with high semantic information. For GAN-based methods, HistoGAN \cite{xue2021selective} proposes a conditional GAN model used for synthesizing realistic histopathology image patches conditioned on class labels. CycleGAN \cite{wei2019generative}	proposes an image translation model that generates synthetic images of adenomatous colorectal polyps and proposes a filtration module called Path-Rank-Filter that enhances the presence of adenomatous features in generated images.

However, GAN-based methods require a separate training process and cannot produce stable results while none of the above current mixing-based methods has fully explored the essential features of the field of pathology. We extend the mixing-based data augmentation techniques based on the following three observations:

1. Domain characteristics of pathology images. 
According to our observation, there are three basic and essential pathology image characteristics in the pathology field.
(1) Local specificity. Factors such as the size of cells, the ratio of the nucleus to the cytoplasm, and the distribution of cell chromatin are important components of pathology image features. They are often scattered in various parts of the image, requiring careful observation of local areas. (2) Global distribution characteristics. Factors such as the relative size of cells, the orientation of cell clusters, and the consistency of cell spacing need to be analyzed as a whole for the entire image. (3) Inner/outer-sample instance relationship. In pathology images, there are different granularities of pathology instances in the same sample, such as cells, tissues, etc. Instances of the same category appear similar, while the relationship of different instances acts as the main feature. 
The relationship between instances in the same image, which is called the \textbf{inner-sample instance relationship}, is a discriminative element in lesion identification. 
Besides, the relationship across instances in different samples is called \textbf{outer-sample instance relationship}. Once we capture the proper outer relationship, we can combine the related instances into a meaningful pathology image. Therefore, we need a proper mixing approach as a data augmentation method that well models the inner-sample relationship and combines instances in proper outer-sample relationships to generate new images.

2. Perturbation information. Pathology images are artificially processed (stained/photographed), which inevitably introduces noise, 
leading to inconsistency in the distribution of pathology images corresponding to the same label. In fact, the attention map in earlier pathology image analysis work always highlights some unrelated regions in the images \cite{zhang2022shuffle}, which also validates the impact of perturbation information. 
Therefore, it is much tougher to do the augmentation while preserving the labeling consistency.

3. Different data distribution between pathology images. 
In the pathology field, due to the difference in collection devices or preprocess methods during the pathology data collection process, a large gap exists between images under the same pathology task.
Thus, previous mixing-based methods will intermingle a variety of different distribution information, 
making the subsequent learning procedure harder.
Therefore, we need a careful training strategy to reduce the difficulty in the learning process.

\begin{figure*}
  \centering
  \includegraphics[width=6.8in]{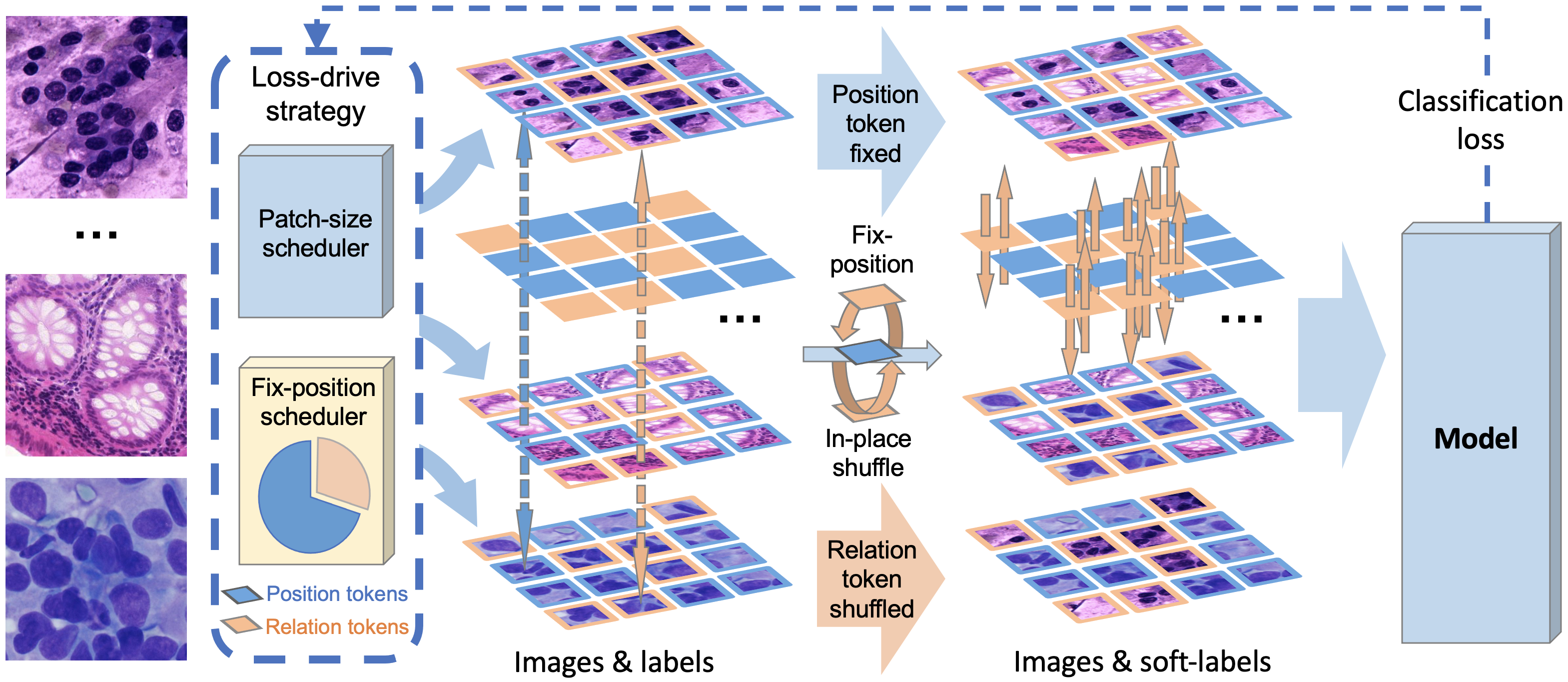}
  \caption{Overview of CellMix containing two components: In-place shuffle strategy and Loss-drive curriculum learning process. First, images are split into patches. Then in Fix-position in-place shuffle strategy, we randomly select position tokens that do not participate in the shuffle process, shown in blue. Yellow patches are relation tokens to be in-place shuffled. Next, in the right part of the figure, we shuffle relation tokens across the batch, remaining their relative positions the same in images. Soft labels are generated from ground truth labels according to the fix-position ratio. For the Loss-drive strategy in the blue box, we designed a Patch-size scheduler and Fix-position scheduler to adaptively guide the learning process according to the model loss. It is introduced specifically in Fig.\ref{fig:LossDrive_Structure}.}
  \label{fig:CellMix_Structure}
\end{figure*}

To address the three issues mentioned above, we propose an instance relationship based augmentation framework based on an in-place shuffle strategy, namely CellMix.
Specifically, we split the images into patches and shuffle them in the same position across one batch at a certain proportion. We use the same proportion to blend the labels of these batch samples. This process is called the in-place shuffle strategy (illustrated in Fig. \ref{fig:CellMix_Structure}). 
The patch size is designed to be consistent with the instance's granularity, which contains local specificity.
In this way, the inner/outer-sample instance relationships and global features in pathology images remain relatively stable compared to other methods. Therefore, we recreate newly labeled pathology samples via the shuffle procedure which complies with the characteristics of pathology images.

In addition, inspired by the idea of curriculum learning 
 \cite{bengio2009curriculum}, we devise a loss-drive strategy during the training process. 
Due to the perturbation noise and domain gap across the pathology images, it is a challenging task to directly use the aforementioned augmented data in the downstream training process. 
Curriculum learning imitates the learning process of human beings. It advocates that the model should learn from the easy samples, and gradually advance to the complex samples. It is proven to be very effective in dealing with complex tasks in numerous research \cite{xu2020curriculum,su2020dialogue,liu2022competence,lu2021exploiting}. 
Therefore, in our CellMix framework, we design a fix-position ratio scheduler and a patch-size scheduler for the in-place shuffle process. We continuously change the patch size and the fix-position ratio of pathology images to reduce the difficulty of the training task. In this way, we can gradually enhance the modeling ability towards biased pathology images and fit the domain gap adaptively. 
Furthermore, to make the training model control the schedule of the learning process by itself, we devise a loss-drive ratio changing guideline. We judge the learning procedure by the loss of our model and alter the difficulty respectively, which is also a crucial approach to further improve the training efficiency.

In summary, our contribution can be concluded as the following four aspects:

1. We are one of the first to develop data augmentation in pathology image classification and summarize three key issues in the augmentation process. These issues include the characteristics in the pathology field, especially the relationship modeling and original noise in pathology images, which should be highly considered in the augmentation process.

2. We propose CellMix, which is a plug-and-play and parameter-free framework based on an in-place shuffle strategy, which explicitly considers the characteristics of pathology images. Specially, in-place shuffle strategy are proposed to exploit the inner/outer-sample instance relationship and achieve data augmentation in a mixing-based manner.

3. 
We devise a loss-drive curriculum learning strategy
during training, making the training process adaptively to the augmented data and exploring the various instance relationships.

4. The experiments conclude that our CellMix framework outperforms the previous SOTA method among 7 different histopathology and cytopathology datasets.

\section{Related Works}
\subsection{Data Augmentation}

Data augmentation has been widely used in deep learning models. Mixup \cite{zhang2017mixup} produces an augmented image by a pixel-wise weighted combination of two images. Cutout \cite{devries2017improved} proposes to mask fixed-size square regions of the input training images. CutMix \cite{yun2019cutmix} randomly masks a rectangular-shaped region of one image and replaces it with the corresponding position of another image. Patchup \cite{faramarzi2020patchup} and Manifold mixup \cite{verma2019manifold} extend the concept of mixup from input space to hidden feature space. FMix \cite{harris2020fmix} uses random-shape masks sampled from Fourier space. PuzzleMix \cite{kim2020puzzle}, SaliencyMix \cite{uddin2020saliencymix} and Co-Mix \cite{kim2021co} focus on image saliency analysis. The mask tries to reveal the most salient regions of images and maximize the saliency of the augmented image. ResizeMix \cite{qin2020resizemix} directly resizes the source image to a small patch and pastes it on another image. RandomMix \cite{liu2022randommix} randomly selects a mixing sample data augmentation method from candidates, which increases the diversity of the mixed samples.

\subsection{Curriculum Learning}

Bengio et al. \cite{bengio2009curriculum} proposed a new learning paradigm called curriculum learning (CL). In this strategy, a model is learned by gradually including easy to complex samples in training in order to increase the entropy of training samples. Previous research have proved that CL is consistent with the principle in human teaching \cite{khan2011humans, basu2013teaching}. 
It often utilizes prior knowledge provided by instructors as guidance for curriculum design. It means that the predetermined curriculum heavily relies on the quality of prior knowledge while ignoring feedback about the learner.

To alleviate this issue, Kumar et al. \cite{kumar2010self} designed a new formulation, called self-paced learning (SPL). SPL adds a regularization term into the learning objective as the curriculum design. SPL leaves learners the freedom to adjust to the actual curriculum according to their learning paces.
Various SPL-based applications have been proposed recently \cite{jiang2014easy, tang2012shifting, kumar2011learning}.  However, SPL is unable to deal with prior knowledge. Self-paced curriculum learning (SPCL) \cite{jiang2015self}, as an “instructor-student-collaborative” learning model, considers both prior knowledge and information learned during training, which gives inspiration to CellMix.

\section{Method}
\label{sec:Method}

In our CellMix framework for pathology image augmentation, there are two main components (demonstrated in Fig. \ref{fig:CellMix_Structure}). 
First is the in-place shuffle strategy, which regroups the patches in pathology images and generates augmented samples. Second is the
loss-drive strategy,
which includes a patch-size scheduler and a fix-position ratio scheduler driven by the model loss.
This designation in curriculum learning aims to make the training process more effective and is flexible to deal with the pathology noise aforementioned. Besides, with the shift of patch-size and fix-position ratio, different scales of pathology image features are included. In this way, the curriculum learning process also makes the augmented pathology samples more representative.
We will give a detailed description of the two main components in the following sections.

\subsection{In-place Shuffle Strategy}

The in-place shuffle strategy draws lessons from the idea of a mixing-based augmentation approach from the computer vision community.  It holds the hypothesis that image labels preserve consistent visual information which can be mixed to generate interpolative samples. Besides, our in-place shuffle strategy highly considers the characteristics of the pathology field. Here we will introduce the main process of the in-place shuffle process.

Let $I \in \mathbb{R}^{W\times{H}\times{C}}$ represent a pathology image, where $W, H, C$ denote the width, height, and channel of the images respectively. Same with ViT, we first equally split the image into $n$ patches, thus the image can be represented as
\begin{equation}
I = \{ P_0, P_1,\dots, P_n \}, \quad P_i \in \mathbb{R}^{[p,p,C]}
\end{equation}
where $P_i$ denotes each patch, and we assign $p$ to denote the patch size of $P_i$. Let $B$ denote the batch in the training process. In our in-place shuffle strategy, we take $\beta$ as the fix-position ratio. We randomly select $m$ patches in the same position across one batch $B$:

\begin{equation}
m = n\times{\beta},\quad \beta \in [0, 1]
\end{equation}
which are regarded as position tokens $F$ that do not participate in the shuffle process. And other patches are relation tokens $R$, acting as patches to be shuffled. 
By fixing a subset of patches unchanged, we preserve this part of the inner-sample instance relationship, which is essential in pathology modeling.

\begin{figure}[t]
  \centering
  \includegraphics[width=3.3in]{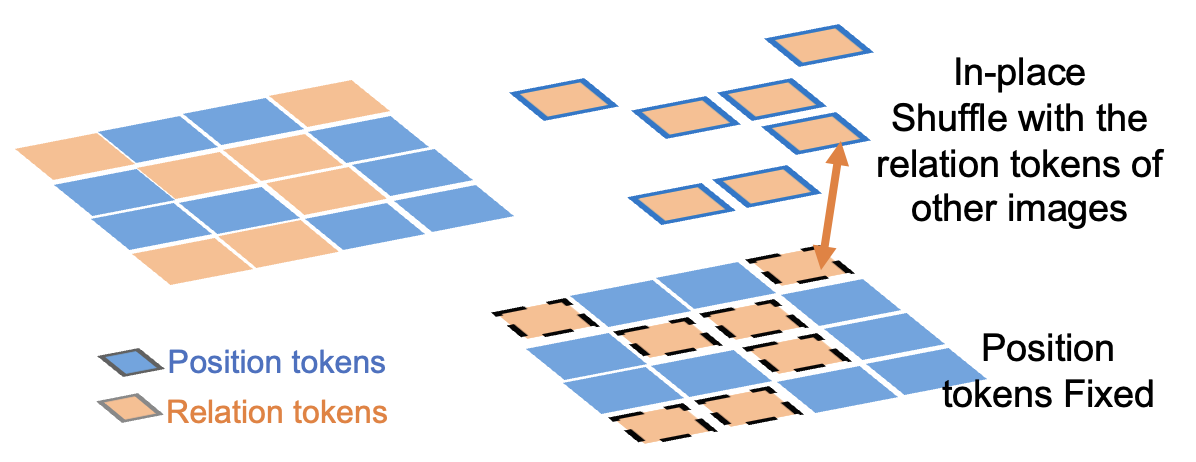}
  \caption{The process of in-place shuffle strategy. The blue tokens represent position tokens while the yellow tokens represent relation tokens. In the in-place shuffle process, we keep the position tokens within the image fixed while shuffle the relation tokens of other images. }
  \label{fig:method_inplace_shuffle}
\end{figure}

Next, we do the in-place shuffle process. As is shown in \ref{fig:method_inplace_shuffle}, we shuffle the relation tokens across each batch $B$ while keeping the position token fixed (illustrated in the right part of Fig. \ref{fig:CellMix_Structure}) which can be formulated as:
\begin{equation}
P_{a, i} = M \odot P_{F, i} + (1 - M) \odot P_{R, i},\quad i \in n
\end{equation}
where $\odot$ is element-wise multiplication and $M$ denotes a binary mask, serving to assign the two kinds of tokens.
\begin{equation}
M = \begin{cases} 0, i \in R\\ 1, i \in F \end{cases}
\end{equation}

To provide a more intuitive understanding, consider the dataset's s-th image, denoted as $I_s$, which consists of 
\begin{equation} I_{s} = \{P_{s,1}, P_{s,2}, P_{s,3}, ..., P_{s,k}, P_{s,k+1}, P_{s,n-1}, P_{s,n} \}.  
\end{equation}

According to the fix-position ratio, a portion of these tokens is randomly assigned as fix-position tokens, while the rest are relation tokens:

\begin{equation} I_{s} = \{F_{s,1}, F_{s,2}, F_{s,3}, ..., R_{s,k}, F_{s,k+1}, R_{s,n-1}, F_{s,n} \}.  
\end{equation}

Through the in-place shuffle process, the portion of relation tokens is replaced by the corresponding tokens from the t-th image, resulting in a transformed image denoted as $I_{a,j}$ in the augmented sample set $I_a$:

\begin{equation} I_{a,j} = \{F_{s,1}, R_{j,2}, F_{s,3}, ..., R_{j,k}, F_{s,k+1}, R_{j,n-1}, F_{s,n} \}.  
\end{equation}

In this way, we augment new pathology sample through the shuffle process.

Meanwhile, the image labels need to be processed synchronously according to the fix-position ratio $f$. Considering that the image label is discrete, We need to use an interpolation method to derive the soft label of the enhanced sample:
\begin{equation}
y_a = f y_f + (1 - f)y_r,\quad y_a \in R^{[cls]}, \quad y_f, y_r \in I^{[cls]}
\end{equation}
where $f \in [0,1]$ denotes the mix-proportion, $y_a$ denotes the soft label of augmented samples, $y_f$,$y_r$ are one-hot vectors. $y_f$ denotes the label of fix-position tokens and $y_r$ denotes the relation tokens. The $[cls]$ denotes the number of classification categories.

With our in-place shuffle strategy, the absolute relationship of the tokens is remained, making the embedding of features fixed in the training process. In this way, compared to other methods, the important relative relationships and global features in pathology images remain stable. Therefore, we recreate pseudo samples via the shuffle procedure which complies with the characteristics of pathology images. 


\begin{figure}[t]
  \centering
  \includegraphics[width=3.3in]{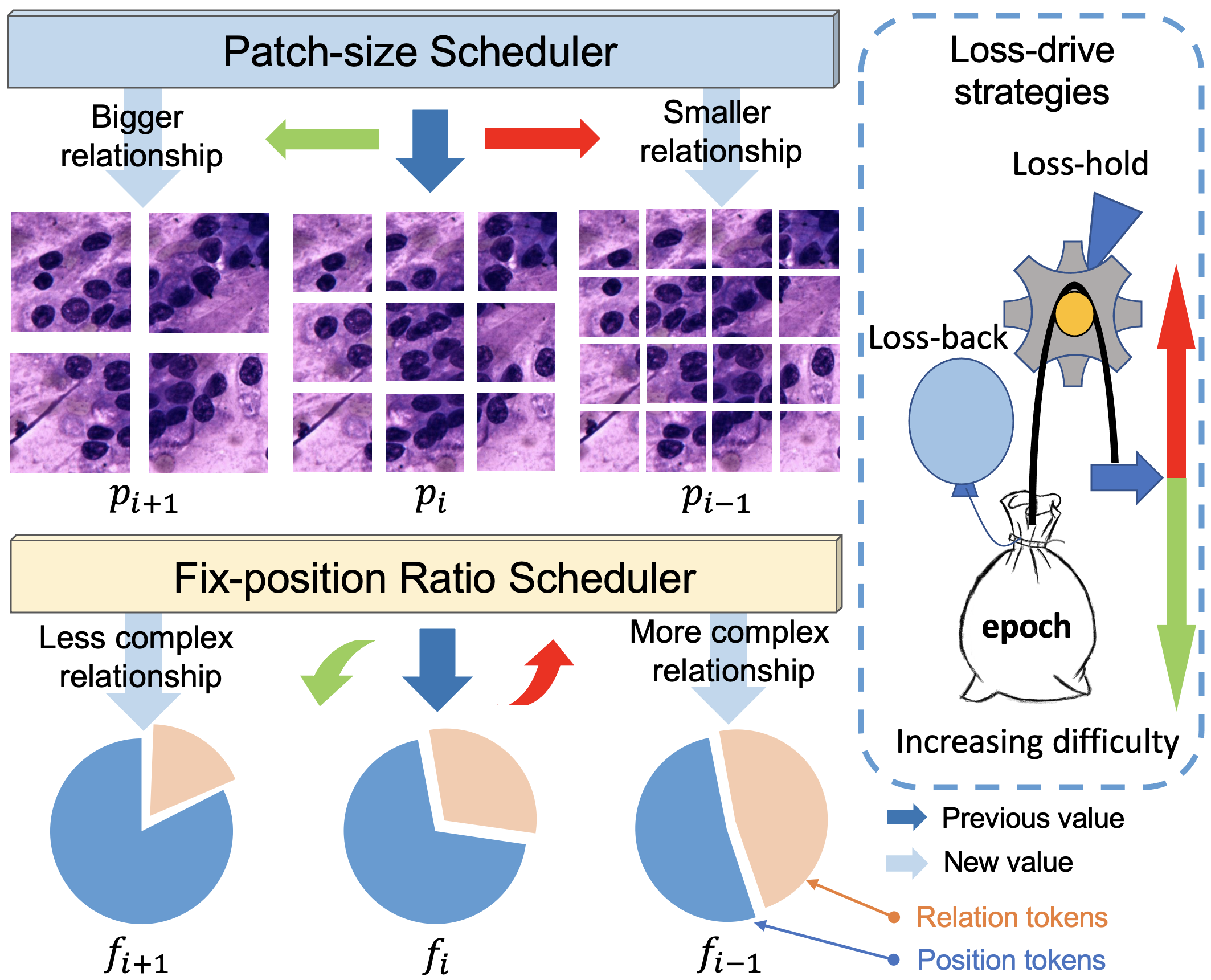}
  \caption{Structure of Loss-drive curriculum learning process. First, we design two schedulers: a Patch-size scheduler and a Fix-position scheduler. The former one schedule patch size from bigger to smaller, while the latter one gradually reduces the ratio of Fix-position tokens (patches) in the shuffle process. Then, Loss-drive strategies, shown in the blue box, are proposed to determine how schedulers work. When the model learns well, we increase the difficulty as the red arrow. Otherwise, with the green arrow, the difficulty decreases. The curriculum remains the same in Loss-hold or steps backward to a simpler one in Loss-back.}
  \label{fig:LossDrive_Structure}
\end{figure}


\subsection{Loss-drive Curriculum Learning}

Due to the influence of perturbation information of the pathology image and the inconsistency of data distribution, it is often difficult to directly use augmented samples for training downstream tasks. Inspired by the idea of curriculum learning,  we imitate the process of human learning knowledge and design training courses from easy to difficult for the model.

\subsubsection{Two Schedulers in Curriculum Learning}

Specifically, in the in-place shuffle strategy, different patch sizes and fix-position ratios will affect the complexity of generated samples and bring about different training difficulties. At the same time, since pathology images actually contain features of different levels and scales, we also hope that the samples augmented by the shuffle process can be representative and flexible. In our framework, we design a patch-size scheduler and fix-position ratio scheduler, as shown in Fig. \ref{fig:LossDrive_Structure}.

With the patch-size scheduler, we continuously reduce the patch size by splitting the pathology images as follows:
\begin{equation}
[p_n, p_{n-1}, \dots, p_1], \quad p_n > p_{n-1} > \dots > p_1
\end{equation}

During this process, we have more and more refined modeling of the relative relationship among instances. From the big-scale relationship to the small-scale relationship, the augmented samples cover the relationship of different granularity.

In the fix-position ratio scheduler, we continuously reduce the fix-position ratio in the shuffle process as follows:
\begin{equation}
    [f_n, f_{n-1}, \dots, f_1]
\end{equation}
where 
\begin{equation}
f_n > f_{n-1} > \dots > f_1, \quad f_i \in (0,1)
\end{equation}

In this way, the mixing efficiency of the in-place shuffle is constantly improved, and the pathological relationship is becoming more and more complex, making the model adaptively enhance the modeling ability. 

Furthermore, in our curriculum learning, we do not learn the subset of total samples every epoch, but all samples with different patch sizes and fix-position ratios. In this way, we manage to plan the augmented pathology images to be trained for the model from easy to difficult. This process eliminates the influence of perturbations in the pathology field and improves the training effect and learning efficiency of augmented samples. At the same time, due to the change of schedule, the model can perceive the pathological characteristics across different scales. This training strategy makes the augmented samples obtained by the in-place shuffle process become more representative. 

\subsubsection{Loss-drive Stragtegy}

On the basis of the above patch size schedule and fix-position ratio schedule, we draw on the idea of self-paced curriculum learning and hope that the training model can adjust the schedule itself to adaptively control the learning process. To this end, we guide the learning process according to the model loss, as shown in the right half of Fig. \ref{fig:LossDrive_Structure}.

Specifically, in the training process, let the loss of the current iteration be $l$, and the learning threshold is $T$. When the loss value of the model is less than a certain threshold, the model has been learned well enough for the current course, which indicates the current course is simple enough for the model. Therefore, we increase the difficulty of the curriculum by reducing the patch size and diminishing the fix-position ratio.
\begin{equation}
p_i \to p_{i-1}; f_i \to f_{i-1} \quad iff, l < T
\end{equation}

On the contrary, when the loss $l$ is larger than the threshold $T$, we give two strategies to control the schedule. First, the loss-hold strategy, which make the schedule remain unchanged.
\begin{equation}
p_i \to p_{i}; f_i \to f_{i} \quad iff, l \geq T
\end{equation}

Second, the loss-back strategy, which pushes back to the previous schedule.

\begin{equation}
p_i \to p_{i+1}; f_i \to f_{i+1} \quad iff, l \geq T
\end{equation}

We will compare the two strategies in the experiments in Section \ref{sec:Experiment}. With the loss-drive strategy, the model can adjust the difficulty by itself, so as to further improve the training efficiency of augmented samples.


\section{Experiment}
\label{sec:Experiment}

In this section, we mainly demonstrate the effectiveness, generalizability, and robustness of CellMix. To evaluate the performance of CellMix, we apply it to 7 different pathology image classification datasets: ROSE \cite{zhang2022shuffle}, pRCC \cite{gao2021instance}, WBC \cite{kouzehkanan2022large}, MARS \footnote{https://www.marsbigdata.com/competition/details?id=21078355578880}, GS \cite{kouzehkanan2022large}, Warwick \cite{sirinukunwattana2017gland} and NCT \cite{kather2018100}. The datasets are introduced in Section \ref{sec:Dataset}, covering different feature scales, dataset scales, and a variety of pathology tasks. As an instance regrouping strategy, CellMix is compared with several mixing-based SOTA data augmentation variants in Section \ref{sec:methods comparison}. Designed for patch-based learning approaches, CellMix also verifies the generalizability onto a wide range of backbones (CNN-based, ViT-based, and Hybrid) in Section \ref{sec:Generalizability}. Different CellMix strategies are introduced in Section \ref{sec:CellMix variants} to explore the design concepts. Interestingly, we examine the effect of curriculum design and feature scale with different prior knowledge in Section \ref{sec:Curriculum Variants}. Lastly, for better interpretability, we use Grad-CAM \cite{selvaraju2017grad} to illustrate that CellMix obtains the most accurate attention area among models in Section \ref{sec:CAM Analysis}.

\subsection{Datasets}
\label{sec:Dataset}


\begin{table}[!t]
    \begin{center}
    \caption{Dataset description.}
    \label{tab: Dataset description}
\resizebox{1\columnwidth}{!}
    {
    \begin{tabular}{c|c|c|c|c}
        \hline
         \textbf{Dataset} & \textbf{Label Nums} &
         \textbf{Images Nums} &
         \textbf{Organ/Tissue} &	 \textbf{Field}\\  
        \hline
         ROSE &	2 &	5088 &	Pancreas &	Cytopathology\\
        pRCC &	2 &	1417 &	Kidney &	Histopathology\\
        WBC &	5 &	14514 &	Blood &	Cytopathology\\
        MARS &	3 &	1770 &	Stomach &	Histopathology \\
        GS & 2 &	700 &	Stomach	& Histopathology\\
        Warwick	& 2	& 165 &	Colon \& Rectum &	Histopathology\\
        NCT &	9 &	100000 &	Colon\& Rectum &	Histopathology\\
        \hline
    \end{tabular}
    }
    \end{center}

\end{table}

The overview of the datasets can be found in Table \ref{tab: Dataset description}.
We cover a wide range of pathology datasets. The ROSE dataset is a cytopathological image dataset of pancreatic tissues, including 1773 pancreatic cancer images and 3315 normal pancreatic cell images. The pRCC dataset is a kidney cancer subtyping dataset, containing 870 types \uppercase\expandafter{\romannumeral1} ROIs and 547 types \uppercase\expandafter{\romannumeral2} ROIs in $2000\times2000$ size. Type \uppercase\expandafter{\romannumeral1} of the pRCC dataset is composed of papillae covered with a single layer of small cells and scant clear or pale cytoplasm and uniform nuclei with inconspicuous nucleoli lying near the basement membrane. Type \uppercase\expandafter{\romannumeral2} of the pRCC dataset is composed of tumor cells with voluminous cytoplasm and pseudostratified high-grade nuclei with prominent nucleoli.
The WBC dataset includes 301 basophil images, 1066 eosinophil images, 3461 lymphocyte images, 795 monocyte images, and 8891 neutrophil images.
The MARS dataset contains 1770 gastric cancer pathological images from the first round of the SEED challenge, including 574 normal gastric pathological images, 403 typical tubular adenocarcinoma pathological images, and 793 typical mucinous adenocarcinoma pathological images. The SEED-Gastric Cancer dataset contains 140 benign gastric pathological images and 560 malignant gastric pathological images from the SEED challenge. The Warwick QU dataset includes 74 benign colorectal tumor cell images and 91 malignant colorectal cancer cell images. The NCT-CRC-HE-100K dataset is a set of 100000 image patches from H$\&$E stained histological images of colorectal cancer and normal tissue, 
including 10407 adipose (ADI), 10566 background (BACK), 11512 debris (DEB), 11557 lymphocytes (LYM), 8896 mucus (MUC), 13536 smooth muscle (MUS), 8763 normal colon mucosa (NORM), 10446 cancer-associated stroma (STR) and 14317 colorectal adenocarcinoma epithelium (TUM) histological images. These datasets are denoted as ROSE, pRCC, WBC, MARS, GS, Warwick, and NCT in this paper without special notification. The overview of datasets refers to Table 1.
\subsection{Implementation Details}
To thoroughly analysis the proposed data-augmentation strategy CellMix, the experiments are fairly done with the same Ubuntu20.04 server (Intel(R) Xeon(R) Platinum 8255C CPU, 512 GB RAM and 8 Nvidia RTX3090 GPU). The PyTorch version is 1.10.0, the CUDA version is 11.3 and the Python version is 3.8. In each experiment, only one GPU is used. The model backbones are built with TIMM, and the backbones are not included, they are built following the official codes.

All 7 datasets are randomly separated into training, validating, and testing sets following a ratio of 7:1:2. As each model converges within 50 epochs, we train each experiment with 50 epochs and take the model with the best (validation) performance as the training output. The final results are reported based on the performance of the independent testing set.

During the experiments, we implement two traditional data augmentation strategies on different datasets. The ROSE, MARS, and pRCC datasets are preprocessed with the first set of pre-processing-based data-augmentation strategies. In the training process, the random rotation and center-cut with a size of  $700\times700$  pixels operations are applied to the input images. Then the input data is randomly horizontally flipped, vertically flipped, resized and color jittered (with the setting of brightness=0.15, contrast=0.3, saturation=0.3, and hue=0.06). In the validating and testing processes, the $700\times700$  pixel center area of each image is cropped and resized without additional operations. Due to the size of the images, we introduce another set of pre-processing data-augmentation strategies on Warwick, GS, WBC, and NCT datasets. Different from the first one, this strategy replaces the random rotation and center-cut operation with resizing operation in the training process. In the validating and testing processes, the second strategy only has the resize operation.

In every training process, a batch size of 8 and a random data augmentation trigging chance of 50$\%$ is used. The hyperparameters of the counterpart methods are set following the official recommendations. The Adam optimizer and cosine weight decay strategy are used in all experiments. The learning rate and the times of learning rate decay for each dataset are listed in Table \ref{tab: Dataset Implementation}.

\subsection{Comparison with SOTA Methods}
\label{sec:methods comparison}

In this section, we provide a comprehensive comparison with many SOTA mixing-based augmentation methods. 
We reproduce the SOTA data augmentation methods \cite{zhang2017mixup, devries2017improved,yun2019cutmix,uddin2020saliencymix,harris2020fmix, qin2020resizemix} with official codes and settings and compare them with the CellMix method.  The baseline does not use any mixing-based data augmentation methods. We use the ViT-base model as the backbone and denote the experiments as Baseline, CutMix, Cutout, FMix, Mixup, ResizeMix, SaliencyMix, and CellMix in the experiment tables. Every online augmentation method is triggered by a 50$\%$ chance in the training process. 


\begin{table*}[!t]
    \begin{center}
    \caption{Accuracy and F1-score comparison with SOTA mixing-based data augmentation methods on datasets including ROSE, pRCC, WBC, GS, Warwick, MARS, and NCT. All counterpart methods are evaluated with ViT-base.}
    \label{tab: methods comparison}
    \tiny
\resizebox{2\columnwidth}{!}
    {
    \begin{tabular}{c|cc|cc|cc|cc|cc|cc|cc}
        \hline
         \textbf{Method} & \multicolumn{2}{|c|}{\textbf{ROSE}} &
         \multicolumn{2}{|c|}{\textbf{pRCC}} &
         \multicolumn{2}{c|}{\textbf{WBC}} &
         \multicolumn{2}{c|}{\textbf{MARS}} &
         \multicolumn{2}{c|}{\textbf{GS}} &
         \multicolumn{2}{c|}{\textbf{Warwick}} &
         \multicolumn{2}{c}{\textbf{NCT}} \\
          & Acc & F1 & Acc & F1 & Acc & F1 & Acc & F1 & Acc & F1 & Acc & F1 & Acc & F1\\    
        \hline
         Baseline & 91.63 &	87.66 &	92.58 &	90.05 &	98.39 &	99.24 &	96.31 &	95.85 &	99.29 &	99.55 &	100.00 &	100.00 &	98.19 &	96.38\\
        CutMix &	91.93 &	88.15 &	95.41 &	93.66 &	98.80 &	99.45 &	96.59 &	96.25 &	100.00 &	100.00 &	100.00 &	100.00 &	99.61 &	98.83\\
         Cutout & 92.72 &	89.58 &	93.29 &	91.24 &	97.60 &	98.44 &	96.88 &	96.49 &	98.57 &	99.11 &	100.00 &	100.00 &	98.33 &	96.11\\
         FMix & 93.41 &	90.41 &	93.64 &	91.35 &	98.32 &	98.99 &	96.31 &	95.79 &	100.00 &	100.00 &	100.00 &	100.00 &	99.39 &	98.59\\
         Mixup & 93.80 &	90.94 &	93.29 &	90.55 &	97.42 &	98.44 &	96.88 &	96.55 &	100.00 &	100.00 &	97.50 &	97.62 &	99.46 &	98.76\\
         ResizeMix & 92.52 &	88.66 &	95.76 &	94.34 &	98.20 &	98.90 &	97.44 &	97.12 &	100.00 &	100.00 &	100.00 &	100.00 &	99.57 &	98.78\\
         SaliencyMix & 93.21 &	90.18 &	94.70 &	93.02 &	98.92 &	99.55 &	98.01 &	97.78 &	100.00 &	100.00 &	100.00 &	100.00 &	99.50 &	98.76\\
         CellMix & \textbf{94.49} & \textbf{92.07} &	 \textbf{97.17}	& \textbf{96.19} & \textbf{99.26}	& \textbf{99.70} & \textbf{98.86}	& \textbf{98.74}	& \textbf{100.00}	& \textbf{100.00}	& \textbf{100.00}	& \textbf{100.00} & \textbf{99.63}	& \textbf{99.12}				\\
        \hline
    \end{tabular}
    }
    \end{center}

\end{table*}


Table \ref{tab: methods comparison} shows CellMix significantly outperforms all other mixing-based augmentation methods. On small-size datasets including GS and Warwick which are relatively simple, the majority of data augmentation methods perform well including CellMix. On medium-size datasets including ROSE, pRCC, WBC and MARS, CellMix yields a striking performance advancement up to 4.59$\%$ in accuracy and 6.14$\%$ in F1-score compared with Baseline, which largely outperforms other SOTA methods.  In addition, CellMix can steadily boost the performance on large-size datasets such as NCT, by 1.47$\%$ for accuracy and 2.74$\%$ for F1-score.

\subsection{Generalizability}
\label{sec:Generalizability}

To verify the generalizability of CellMix, we examine various baseline models divided into three sets: CNN-based backbones, ViT-based backbones and Hybrid backbones. CNN-based backbones include VGG16, VGG19 \cite{simonyan2014very}, Resnet50 \cite{he2016deep}, Xception \cite{chollet2017xception} and Mobilenetv3 \cite{howard2017mobilenets}. ViT-based models, also considered as patch-learning-based models are ViT-base \cite{dosovitskiy2020image} and Swin-base \cite{liu2021swin}. Conformer \cite{gulati2020conformer}, Crossformer \cite{wang2021crossformer} and ResNet50-ViT \cite{dosovitskiy2020image} are Hybrid backbones. Unless specified otherwise, the model implementation is based on official codes and minimal changes are made to hyperparameters of these backbones.


\begin{table*}[!t]     
    \begin{center}
    \caption{ CellMix can steadily boost a wide range of model variants including CNN-based, ViT-based, and Hybrid models on all classification benchmarks. Note that all hyperparameters within one dataset remain the same as Section \ref{tab: methods comparison}.}
    \label{tab: generalizability}
\resizebox{2\columnwidth}{!}
    {
    \begin{tabular}{c|cc|cc|cc|cc|cc|cc|cc}
        \hline
         \textbf{Model} & \multicolumn{2}{|c|}{\textbf{ROSE}} &
         \multicolumn{2}{|c|}{\textbf{pRCC}} &
         \multicolumn{2}{c|}{\textbf{WBC}} &
         \multicolumn{2}{c|}{\textbf{MARS}} &
         \multicolumn{2}{c|}{\textbf{GS}} &
         \multicolumn{2}{c|}{\textbf{Warwick}} &
         \multicolumn{2}{c}{\textbf{NCT}} \\
          & Baseline & CellMix & Baseline & CellMix & Baseline & CellMix & Baseline & CellMix & Baseline & CellMix & Baseline & CellMix & Baseline & CellMix\\    
        \hline
         VGG16 &
	91.44 &	90.85 &	86.22 &	\textbf{86.22} &	98.36 &	\textbf{99.03} &	96.59 &	\textbf{98.01} &	99.29 &	\textbf{99.29} &	96.25 &	\textbf{97.50} & 98.06 &	\textbf{99.55}\\
         VGG19 &
         89.96 &	\textbf{91.63} &	88.34 &	87.28 &	98.25 &	\textbf{98.94} &	96.88 &	\textbf{97.73} &	99.29 &	98.57 &	95.00 &	\textbf{97.50} & 98.10 &	\textbf{99.55}\\
         Resnet50 &
         91.14 &	\textbf{91.24} &	90.46 &	89.40 &	98.92 &	98.50 &	97.44 &	\textbf{97.73} &	100.00 &	\textbf{100.00} &	100.00 &	\textbf{100.00} & 99.19	& \textbf{99.67}\\
         Xception &
         90.65 &	\textbf{90.85} &	82.69 &	\textbf{85.87} &	98.20 &	\textbf{98.46} &	97.44 &	96.59 &	96.43 &	\textbf{98.57} &	100.00 &	\textbf{100.00} & 99.23 &	\textbf{99.64}\\
         Mobilenetv3 &
         87.30 &	86.02 &	75.27 &	71.38 &	97.21 &	\textbf{97.74} &	96.02 &	\textbf{96.02} &	96.43 &	\textbf{96.43} &	95.00 &	93.75 & 98.87 &	\textbf{99.24}\\
         \hline
         ViT-base &
         91.63 &	\textbf{94.49} &	92.58 &	\textbf{97.17} &	98.39 &	\textbf{99.26} &	96.31 &	\textbf{98.86} &	99.29 &	\textbf{100.00} &	100.00 &	\textbf{100.00} & 98.19 &	\textbf{99.63}\\
         Swin-base &
         92.22 &	\textbf{93.31} &	92.23 &	\textbf{92.93} &	98.89 &	98.36 &	96.59 &	\textbf{97.44} &	97.86 &	\textbf{99.29} &	100.00 &	\textbf{100.00} & 97.20 &	\textbf{99.46}\\
         \hline
         Conformer &
         91.34 &	\textbf{92.62} &	92.93 &	90.11 &	97.74 &	\textbf{98.43} &	96.02 &	\textbf{97.44} &	100.00 &	\textbf{100.00} &	100.00 &	\textbf{100.00} & 99.60 &	\textbf{99.60}\\
         Crossformer &
         91.73 &	\textbf{92.91} &	89.40 &	\textbf{89.40} &	97.14 &	\textbf{98.39} &	96.31 &	\textbf{97.44} &	99.29 &	\textbf{99.29} &	97.50 &	\textbf{100.00} & 97.77 &	\textbf{99.43}\\
         ResNet50-ViT &
	92.62 &	\textbf{92.72} &	90.81 &	90.11 &	99.06 &	98.92 &	96.59 &	\textbf{98.01} &	99.29 &	\textbf{99.29} &	100.00 &	\textbf{100.00} & 99.73 &	\textbf{99.82}\\
        \hline
    \end{tabular}
    }
    \end{center}

\end{table*}


In Table \ref{tab: generalizability}, CellMix reveals great advantages on ViT-based models, with up to 4.59$\%$ and 1.43$\%$ improvements on ViT-base and Swin-base models respectively. CellMix cuts input images into non-overlapping patches, which perfectly encounters the patch-learning-based models without breaking the actual value of tokens. The CellMix introduces an instance (patch) regrouping strategy to enhance the tokens’ relationship modeling in the attention-based approaches. This may explain why CellMix performs especially well on patch-learning-based models. For CNN-based backbones and Hybrid backbones, the proposed method consistently boosts the variants in the range of 0.1$\%$ to 3.18$\%$.

In short, Table \ref{tab: generalizability} shows the considerable generalizability of CellMix, considering the improvements of CellMix on non-patch-learning-based methods, and especially striking performance on patch-learning-based models.

\subsection{Variants on In-place Shuffle Strategy with Loss-Drive}
\label{sec:CellMix variants}

CellMix mainly innovates with the fix-position in-place shuffle strategy and loss-drive method. In this section, we explore variants of both the in-place shuffle strategy and the loss-drive method. In the in-place shuffle strategy, a set of randomly selected patches are shuffled across the batch, remaining the relative positions the same in images. For the first strategy of CellMix-group, those alternative patches are from the same image. For the second strategy CellMix-split, each patch is replaced separately, making extremely complex instance groups. In this way, the augmented images, containing regions from various samples are more biased toward each instance rather than focusing on the same sample relationship of the instances.

As for the loss-drive method, we have two variants denoted as Loss-hold and Loss-back. Loss drive is an “instructor-student-collaborative” learning mode considering both prior knowledges of the curriculum and feedback from learners. The specific curriculum in the next iteration is defined by the model performance of the current epoch. If the model learns well, we increase the difficulty of learning. While if not, the curriculum remains the same in Loss-hold or steps backward to a simpler one in Loss-back. Therefore, the model can learn different feature scales in a more comprehensive way.


\begin{table}[!t]
    \begin{center}
    \caption{Accuracy of variants for in-place shuffle strategy with loss-drive method. Group and Split denote the CellMix-group and CellMix-split strategies respectively. Hold means the proposed Loss-hold strategy, and Back means the Loss-back strategy. }
    \label{tab: CellMix variants}
\resizebox{1\columnwidth}{!}
    {
    \begin{tabular}{c|c|c|c|c|c|c|c}
        \hline
         \textbf{Method} & \textbf{ROSE} &
         \textbf{pRCC} &
         \textbf{WBC} &
         \textbf{MARS} &
         \textbf{GS} &
         \textbf{Warwick} &
         \textbf{NCT} \\  
        \hline
         Baseline &	91.63 &	92.58 &	98.39 &	96.31 &	99.29 &	100.00 & 98.19\\
        Group-Hold &	\textbf{94.49} &	\textbf{97.17} &	\textbf{99.26} &	\textbf{98.86} &	\textbf{100.00} &	\textbf{100.00} & \textbf{99.63}\\
         Group-Back &	93.60 &	96.47 &	99.03 &	98.30 &	100.00 &	100.00 & 99.61\\
         Split-Hold &	92.91 &	93.99 &	98.82 &	97.44 &	100.00 &	100.00 & 99.57\\
         Split-Back &	93.50 &	93.64 &	98.27 &	98.01 &	100.00 &	100.00 & 99.55\\
        \hline
    \end{tabular}
    }
    \end{center}

\end{table}


As shown in Table \ref{tab: CellMix variants}, all variants of CellMix can steadily boost the baseline performance on all benchmarks. And CellMix-group with Loss-hold strategy, as the best variant, lifts the accuracy of ViT up to a remarkable 4.59$\%$. Other variants lift the baseline result up to 3.89$\%$, 1.41$\%$ and 1.87$\%$ for CellMix-group-Loss-back, CellMix-split-Loss-hold and CellMix-split-Loss-back respectively. Although all variants are effective, CellMix-split may over complicate the augmentation images and break the absolute relationship of instances, resulting in negative effects on the performance. Loss-hold is better than Loss-back considering the potential over-simplified curriculum in the Loss-back strategy. CellMix-group with Loss-hold strategy is the default strategy denoted as CellMix unless specified.



\begin{figure*}
  \centering
  \includegraphics[width=6in]{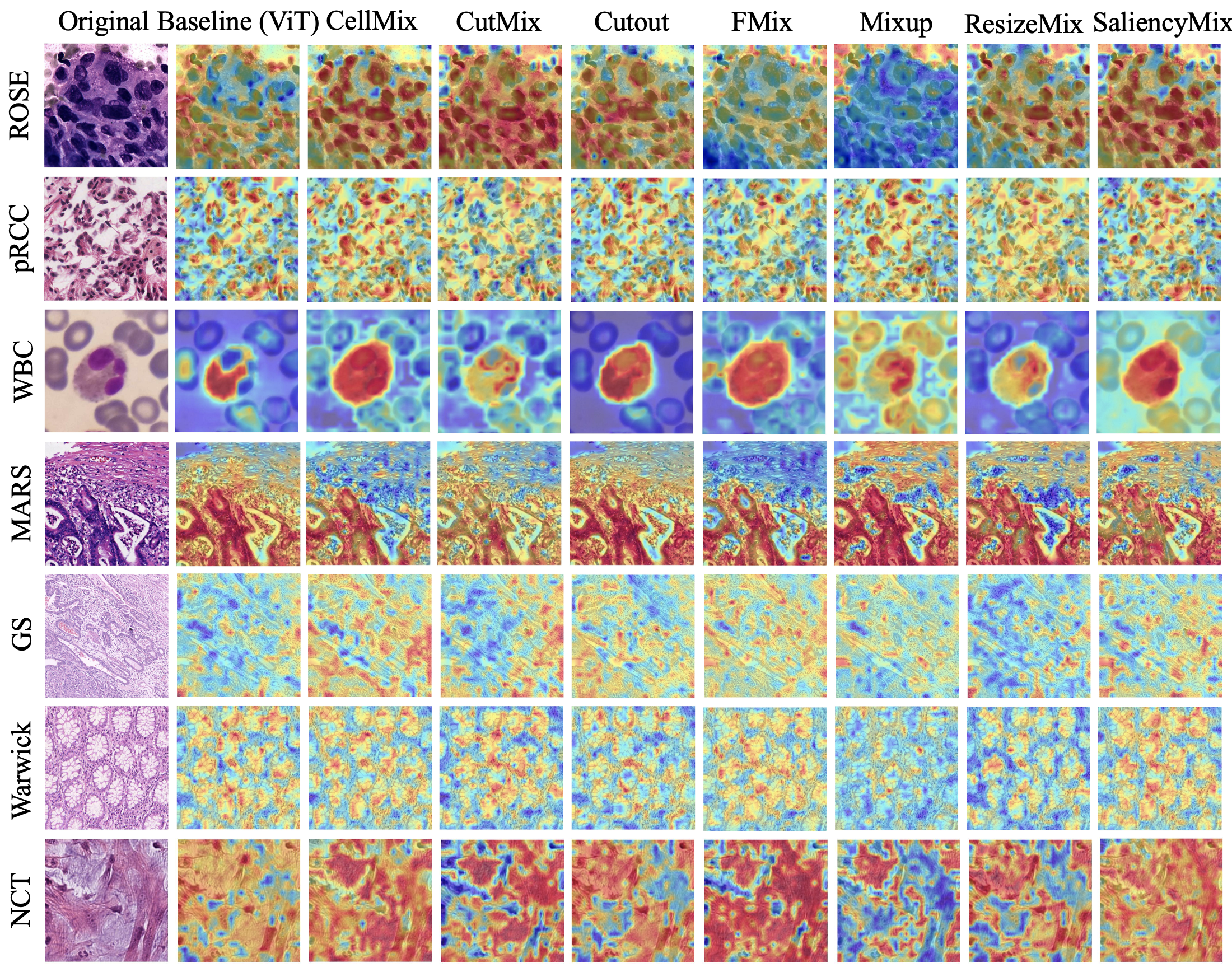}
  \caption{CAM visualizations for trained models on un-augmented images. The first column is the original image from ROSE, pRCC, WBC, MARS, GS, Warwick and NCT, and the other columns are Grad-CAM results. As shown in the figure, CellMix guides the model to precisely focus on target instances and can accurately recognize the contour of the cell area.}
  \label{fig: CAM_unaugmented}
\end{figure*}

\begin{table*}[!t]
    \begin{center}
        \caption{Implementation settings on the datasets.}
        \label{tab: Dataset Implementation}
    \tiny
\scriptsize
    {
    \begin{tabular}{c|c|c|c|c|c|c}
        \hline
         \textbf{Dataset} & \textbf{Learning Rate} &
         \textbf{Cosine Learning Rate Decay} &
         \textbf{Batch Size}  & \textbf{Total Epochs} & \textbf{Image Resize} & \textbf{GPU System}\\  
        \hline
        ROSE &	0.000006 &	to 0.1 times &	8 &	50 &	384	& 3090\\
        pRCC &	0.000002 &	to 0.5 times &	8 &	50 &	384	& 3090\\
        WBC &	0.000005 &	to 0.2 times &	8 &	50 &	384	& 3090\\
        GS &	0.000008 &	to 0.2 times &	8 &	50 &	384	& 3090\\
        Warwick &	0.00001 &	to 0.2 times &	8 &	50 &	384 &	3090\\
        MARS &	0.00002 &	to 0.3 times &	8 &	50 &	384	& 3090\\
        NCT &	0.000007 &	to 0.35 times &	8 &	50 &	384	& A100\\
        \hline
    \end{tabular}
    }
    \end{center}

\end{table*}

\begin{table*}[!t]
    \begin{center}
        \caption{ CellMix can steadily boost the model performance on variety of datasets of both $384\times384$ and $224\times224$ sizes. }
        \label{tab: Image Sizes}
\resizebox{2\columnwidth}{!}
    {
    \begin{tabular}{c|cc|cc|cc|cc|cc|cc|cc}
        \hline
         \textbf{Image Size} & \multicolumn{2}{|c|}{\textbf{ROSE}} &
         \multicolumn{2}{|c|}{\textbf{pRCC}} &
         \multicolumn{2}{c|}{\textbf{WBC}} &
         \multicolumn{2}{c|}{\textbf{MARS}} &
         \multicolumn{2}{c|}{\textbf{GS}} &
         \multicolumn{2}{c|}{\textbf{Warwick}} &
         \multicolumn{2}{c}{\textbf{NCT}} \\
          & Baseline & CellMix & Baseline & CellMix & Baseline & CellMix & Baseline & CellMix & Baseline & CellMix & Baseline & CellMix & Baseline & CellMix\\    
        \hline
         384 &
         91.63 &	\textbf{94.49} &	92.58 &	\textbf{97.17} &	98.39 &	\textbf{99.26} &	96.31 &	\textbf{98.86} &	99.29 &	\textbf{100.00} &	100.00 &	\textbf{100.00} & 98.19 &	\textbf{99.63}\\
         224 &
         91.73 &	\textbf{93.80} &	92.93 &	91.17 &	98.55 &	\textbf{98.57} &	97.44 &	\textbf{98.01} &	100.00 &	\textbf{100.00} &	98.75 &	\textbf{100.00} &	98.07 &	\textbf{99.48}\\

        \hline
    \end{tabular}
    }
    \end{center}

\end{table*}

\subsection{Image Sizes}

As shown in Table \ref{tab: Dataset Implementation}, the image sizes of all datasets are $384\times 384$. We chose $384 \times 384$ because it has more common factors than the common multiples of 16 (token embedding size of Transformer). This enables the fine-grained analysis of instance relationship modeling. In the loss-drive strategy of CellMix, the input images are split into patches according to a given patch size which is continuously decreased. And the minimum patch size is $16\times 16$ consistent with Transformer embedding. To ensure no overlapping or lost information, the size of each patch size is required to be multiple of the minimum value ($16$). $384\times384$ images can be split into more patches with different sizes, compared with $224\times224$ images. In this way, more pairs of instances modeling with various feature scales can be explored.

However, one might wonder whether CellMix still works when images are $224\times224$. We ablate both $384\times384$ and $224\times224$ images on the ViT-base model (in Table \ref{tab: Image Sizes}). CellMix boosts the performance of baseline on the majority of datasets by a maximum of 2.07$\%$. It shows that CellMix makes a striking improvement regardless of image size. In addition, the accuracy of CellMix on $224\times224$ images is lower than its counterpart on $384\times384$ images. It supports that different feature scales contribute to better relationship modeling.

\subsection{Curriculum Variants on Instance Relationship Hyper-parameters}
\label{sec:Curriculum Variants}

As mentioned in Section \ref{sec:Method}, patch size and fix-position ratio of samples represent the difficulty of the curriculum. With an adaptive curriculum, the model can perceive the pathology characteristics across different scales and eliminate the perturbation information of pathology images. The prior knowledge is the range, i.e. learning space, of both two variables. In this section, we ablate different prior knowledge choices to explore the effect of curriculum design and feature scale. We fix different initial values for patch size from 16 to 192 as well as fix position ratio from 0.5 to 0.9, shown in Table \ref{tab: curriculum variants-fix patch size} and Table \ref{tab: Curriculum variants - shuffle ratio} respectively. In addition, we choose different patch size lists including [128, 64, 32, 16] and [196, 96, 48, 16], shown in Table \ref{tab: curriculum variants-fix patch size}.


\begin{table}[!t]
    \begin{center}
    \caption{Curriculum variants for different patch sizes. The upper of the table is fixed in the range of 16 to 192. The lower contains different patch size lists. Full, as the default set, means patch size list [192, 128, 96, 64, 48, 32,16]. Even is [128, 64, 32, 16], while Odd is [196, 96, 48, 16]. The fix-position ratio scheduler remains the same.}
    \label{tab: curriculum variants-fix patch size}
\resizebox{0.5\columnwidth}{!}
    {
    \begin{tabular}{c|c|c|c}
        \hline
         \textbf{Patch Size} & \textbf{ROSE} &
         \textbf{pRCC} &
         \textbf{WBC}  \\  
        \hline
         16	& 93.90	& 94.70	& 98.50\\
        32 &	94.09 &	94.70 &	98.76\\
         48 &	93.50 &	\textbf{96.11} &	\textbf{98.78}\\
         64	& \textbf{94.19}	& 95.05 &	98.71\\
         96 &	94.00	& 95.05 &	98.20\\
         128 &	93.31 &	95.05 &	98.78\\
         192 &	92.81 &	95.05 &	98.27\\
        \hline
        Even &	93.70 &	96.47 &	98.62\\
        Odd &	94.19 &	95.76 &	98.32\\
        Full &	\textbf{94.49} &	\textbf{97.17} &	\textbf{99.26}\\
        \hline
    \end{tabular}
    }
    \end{center}

\end{table}


In the upper part of Table \ref{tab: curriculum variants-fix patch size}, CellMix performs the best with fix-patch size 64 on ROSE and 48 on pRCC and WBC, while too small or too large patch size shows poor performance. In fact, instances (i.e. cells, tissues, etc.) are sparsely distributed in the majority of pathology images. With a large patch size, it is more likely to generate augmented samples with misleading supervisory signals. While too small, instances are split into pieces resulting in the loss of semantic information. This result verifies the capability of instance-based shuffle strategy and supports our hypothesis that local (instance) specificity, such as characteristics of cells, plays an important role in modeling pathology image features.


    

The lower part of Table \ref{tab: curriculum variants-fix patch size} shows CellMix performs the best with a full patch size list. Specifically, it outperforms CellMix without full patch size lists in a range of 0.3$\%$ to 1.41$\%$. Shortening the patch size list reduces learnable feature scales and the accuracy of results, which means a strategy with more comprehensive feature scales can work better in relationship modeling. In Table \ref{tab: Curriculum variants - shuffle ratio}, CellMix with adaptive fix-position ratio outperforms others with given ratios from 0.13$\%$ to 2.82$\%$. The drop in performance of the variant without a full patch size list and the variant without an adaptive fix-position ratio is consistent with the perspective that different feature scales contribute to better relationship modeling.





\subsection{Curriculum Variants on Loss-drive Strategy}

\begin{table}[!t]
    \begin{center}
    \caption{Variants on the fix-position ratio scheduler. Linear Decay is the strategy that the fix-position ratio steadily and rigidly reduces, while Adaptive is the one used in CellMix which considers the actual learning efficiency and dynamically adjust the curriculum.}
    \label{tab: Variants on Fix-position Ratio}
\resizebox{0.6\columnwidth}{!}
    {
    \begin{tabular}{c|c|c|c}
        \hline
         \textbf{Fix-Position Ratio} & \textbf{ROSE} &
         \textbf{pRCC} &
         \textbf{WBC}  \\  
        \hline
        Baseline &	91.63 &	92.58 &	98.39\\
         Linear Decay &	92.72 &	95.05 &	98.92\\
         Adaptive &	\textbf{94.49} &	\textbf{97.17} &	\textbf{99.26}\\
        \hline
    \end{tabular}
    }
    \end{center}

\end{table}

In CellMix, we design a patch size scheduler and a fix-position ratio scheduler to enhance the capability of the loss-drive strategy. Additional ablation experiments are conducted on the patch size scheduler in Section \ref{sec: Variants on Fix-position Ratio} and on the fix-position ratio scheduler in Section \ref{sec: Variants on Patch Size Scheduler}. We also ablate the values of thresholds in loss-drive to explore its effect, shown in Section \ref{sec: Variants on Loss-drive Threshold}.

\subsubsection{Variants on Fix-Position Ratio Scheduler}
\label{sec: Variants on Fix-position Ratio}

Loss-drive strategy adaptively reduces the ratio of fix-position tokens in the shuffle process. When the model learns well, we
decrease the fix-position ratio, while if not, it grows or remains the same. We conduct an ablation experiment on the linear fix-position ratio scheduler (denoted as Linear Decay in Table \ref{tab: Variants on Fix-position Ratio}), in which the fix-position ratio reduces in each iteration regardless of the real learning performance. It is to verify the efficiency of the adaptive fix-position ratio scheduler.


\begin{table}[!t]

    \begin{center}
    \caption{Curriculum variants for given fix-position ratios. For example, 0.9 means 90$\%$ patches of one image are fixed in the in-place shuffle process, with 10$\%$ shuffled. Adaptive, as the default setting, represents the continuously decreasing fix-position ratio.}
    \label{tab: Curriculum variants - shuffle ratio}
\resizebox{0.6\columnwidth}{!}
    {
    \begin{tabular}{c|c|c|c}
        \hline
         \textbf{Fix Position Ratio} & \textbf{ROSE} &
         \textbf{pRCC} &
         \textbf{WBC}  \\  
        \hline
         0.5 &	92.91 &	94.35	& 97.56\\
        0.6 &	93.90 &	94.35 &	98.94\\
         0.7 &	92.13 &	94.70 &	97.35\\
         0.8 &	93.31 &	94.35 &	98.80\\
         0.9 &	93.41 &	94.35 &	99.03\\
         Adaptive &	\textbf{94.49} &	\textbf{97.17} &	\textbf{99.26}\\
        \hline
    \end{tabular}
    }
    \end{center}

\end{table}


Both Linear Decay and Adaptive can boost the performance of baseline model (illustrated in Table \ref{tab: Variants on Fix-position Ratio}). It shows the necessity of a fix-position ratio scheduler. In addition, Adaptive outperformes Linear Decay, with 1.77$\%$, 2.12$\%$ and 0.34$\%$ on ROSE, pRCC and WBC respectively. It shows that leaving learners the freedom to adjust to the actual curriculum according to their learning paces contributes to the improvement of the performance.


\subsubsection{Variants on Patch Size Scheduler}
\label{sec: Variants on Patch Size Scheduler}

Except for adaptively adjusting to the actual curriculum (patch size) in loss-drive, we consider other variants including Reverse, Linear, Random and Loop. For the Linear strategy, we gradually switch the legal patch size from small to big, while in the Reverse strategy, we reverse the process. We randomly assign a patch size in each iteration in the Random strategy. In Loop, we cyclically go through every patch size.


\begin{table}[!t]
    \begin{center}
        \caption{Variants on patch size schedulers. Reverse means patch size is switched from big to small. Linear is from small to big. Random is to randomly assign a patch size in each epoch. Loop means every patch size cycles in a loop. Loss-drive is the same in CellMix.}
        \label{tab: Variants on Patch Size Scheduler}
\scriptsize
    {
    \begin{tabular}{c|c|c|c}
        \hline
         \textbf{Scheduler} & \textbf{ROSE} &
         \textbf{pRCC} &
         \textbf{WBC}  \\  
        \hline
        Baseline &	91.63 &	92.58 &	98.39\\
         Reverse &	93.21 &	96.47 &	98.50\\
         Linear &	93.80 &	95.76 &	98.82\\
         Random	& 93.90 &	95.41 &	98.20\\
         Loop &	92.32	& 95.41 &	98.78\\
         Loss-drive &	\textbf{94.49} &	\textbf{97.17} &	\textbf{99.26}\\
        \hline
    \end{tabular}
    }
    \end{center}

\end{table}

As shown in Table \ref{tab: Variants on Patch Size Scheduler}, all variants boost the baseline performance. With patch size adjusted during the training procedure, the model can perceive the pathology characteristics across different scales. Therefore, Table \ref{tab: Variants on Patch Size Scheduler} shows that different feature scales can facilitate relationship modeling. In addition, the loss-drive strategy is better than other variants as it adapts the curriculum according to the information learned during training.


\subsubsection{Variants on Loss-drive Threshold}
\label{sec: Variants on Loss-drive Threshold}

In the loss-drive strategy, we guide the learning process according to the comparison between the model loss and a certain threshold. For example, when the model loss is less than the threshold, the model is considered to learn well enough. Therefore, we increase the difficulty of the curriculum in the next iteration. Threshold plays an important role in this procedure. We ablate the setting of the threshold in this section, from 2.0 to 8.0, to explore its effect.

CellMix outperformes Baseline with all thresholds. Table \ref{tab: Variants on Loss-drive Threshold} shows that 4.0 is a proper initial value for the threshold. Actually, the value of threshold is closely related to the learning capability of the model, and may be different on various datasets. However, although ROSE, pRCC and WBC are datasets in different pathology domains from cytopathology to histopathology, their initial thresholds are all 4.0, which indicates that the model learns their features at a similar rate, and even in a similar way.

\begin{table}[!t]
    \begin{center}
        \caption{Variants on the loss-drive threshold. The range of the initial threshold is from 2.0 to 8.0. The loss-drive strategy with 4.0 as threshold outperforms other variants.}
        \label{tab: Variants on Loss-drive Threshold}
\scriptsize
    {
    \begin{tabular}{c|c|c|c}
        \hline
         \textbf{Threshold} & \textbf{ROSE} &
         \textbf{pRCC} &
         \textbf{WBC}  \\  
        \hline
        Baseline &	91.63 &	92.58 &	98.39\\
         2.0 &	93.31 &	96.82 &	99.01\\
         4.0 &	\textbf{94.49} &	\textbf{97.17} &	\textbf{99.26}\\
         6.0 &	93.31 &	96.82 &	99.03\\
         8.0 &	94.00 &	96.47 &	99.03\\
        \hline
    \end{tabular}
    }
    \end{center}

\end{table}

\begin{figure*}
  \centering
  \includegraphics[width=6in]{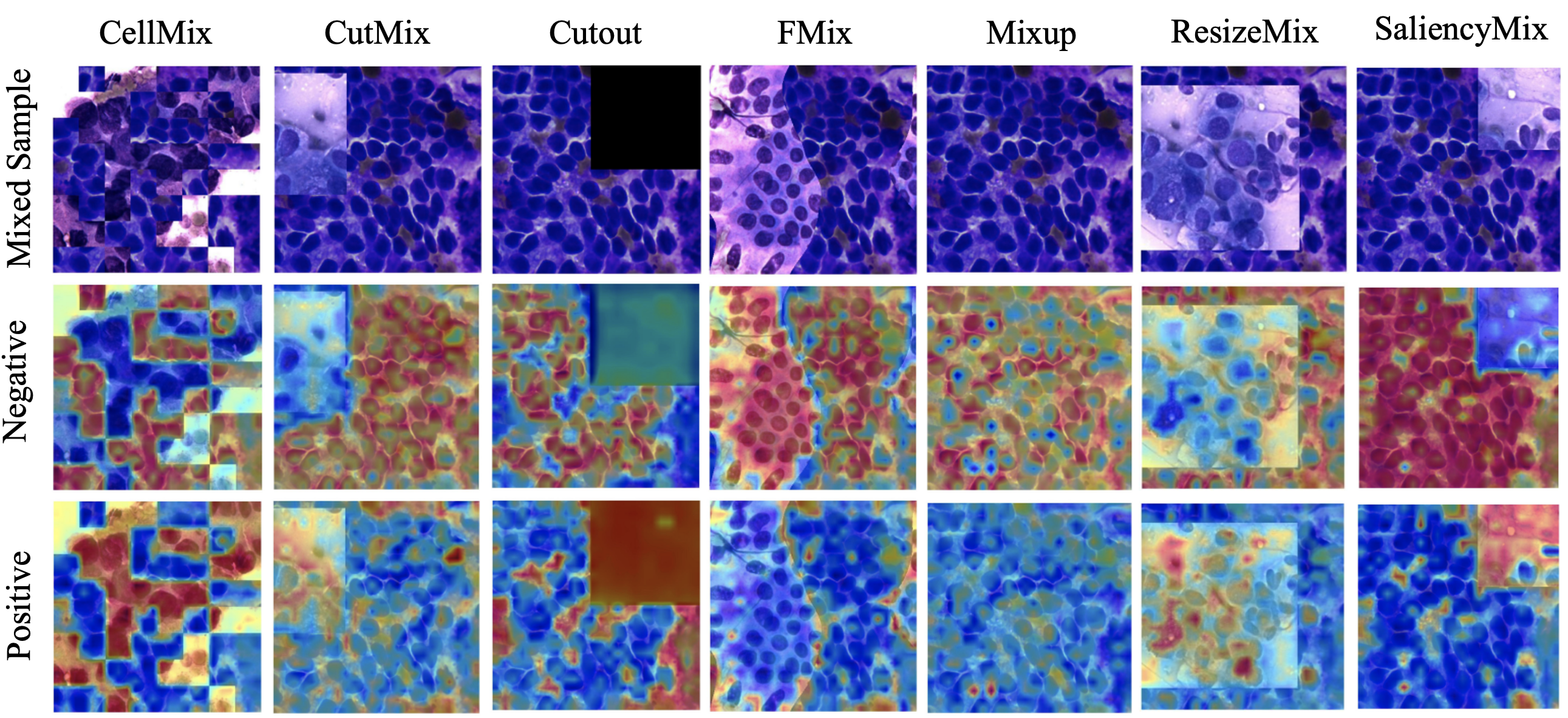}
  \caption{CAM visualization of augmented samples.
  The first row is the mixed images. They are augmented by the corresponding mixing-based methods. The last two rows are Grad-CAM results for negative and positive predictions. CellMix accurately identifies the boundary of patches and instances. It also tells the differences between negative and positive.}
  \label{fig:CAM_augmented_Appendix}
\end{figure*}

\subsection{Random Labels}
We explore the function of labels on augmented images. The soft labels are generated based on the proportion of the mixed patches. One might be wondering if the data augmentation method still works with the wrong labels. To tackle the aforementioned issue, we select different proportions of images and generate random labels for them. Then the dataset is augmented with these random labels and ground truth labels from unselected images.


\begin{table}[!t]
    \begin{center}
    \caption{Accuracy for the dataset with different proportions of random labels. From 0 to 0.8, they denote the ratio of wrong labels. For example, 0.2 denotes 20$\%$ wrong labels with 80$\%$ ground truth. }
    \label{tab: Random Label}
\resizebox{0.6\columnwidth}{!}
    {
    \begin{tabular}{c|c|c|c}
        \hline
         \textbf{Random Ratio} & \textbf{ROSE} &
         \textbf{pRCC} &
         \textbf{WBC}  \\  
        \hline
         0 &	\textbf{94.49} &	\textbf{97.17} &	\textbf{99.26}\\
        0.2 &	54.93 &	66.90 &	39.16\\
         0.4 &	72.68 &	52.67 &	78.84\\
         0.6 &	72.39 &	70.46 &	58.93\\
         0.8 &	53.45 &	50.53 &	20.60\\
        \hline
    \end{tabular}
    }
    \end{center}

\end{table}


Table \ref{tab: Random Label} illustrates the drop in performance when labels of augmented images are wrong. It verifies that soft labels, as guidance during model training, are indeed learned by the model. In addition, it shows that CellMix has no over-fitting issue, otherwise, the results of different random ratios would be similar to each other.

\subsection{CAM Analysis}
\label{sec:CAM Analysis}

We extract Grad-CAM \cite{selvaraju2017grad} for both un-augmented images and augmented images to investigate the regions of the input image where the model focuses to recognize an object.

Fig. \ref{fig: CAM_unaugmented} shows that CellMix guides the model to precisely focus on the target object compared to other methods, presenting solid interpretability in cell identification. It shows that the model fully learns the feature of cell instances (local specificity) and the difference between instances and background (global distribution).

We also visualize Grad-CAM on augmented images in Fig. \ref{fig:CAM_augmented}. It can be seen that CellMix effectively focuses on the corresponding features and precisely localizes the instances (cell, cell cluster, tissue structure, etc.) in the scene. Besides, it can be seen clearly that CellMix pays attention to different cells when making negative and positive predictions, and accurately identifies the boundary of different patches. It proves that CellMix indeed learns the difference between negative and positive instances and the inner/outer-sample instance relationship.

\section{Conclusion}
In conclusion, we introduce CellMix data augmentation method, which aims for the characteristics of pathology image analysis and forces the models to observe the relationship between the instances. 
We propose a distribution-based in-place shuffle strategy that can regroup patches (containing instances) in the image (bag) level, with two kinds of tokens assigned. At the image level, effective pseudo samples force the models to observe the instance relationship. We devise a loss-drive curriculum learning strategy during the training process to cover the multiple potential feature scales and alter the modeling difficulty adaptively.

On various pathology image classification benchmarks, CellMix significantly outperforms all other mixing-based augmentation methods. CellMix boosts the performance of the ViT-base model up to 4.59$\%$ in accuracy and 6.14$\%$ in F1-score. The inspiring SOTA results with various models over 7 datasets prove the effectiveness of the core idea.


\bibliographystyle{IEEEtran}
\bibliography{IEEEabrv,egbib}

\end{document}